\documentclass[11pt,a4paper]{article}
\usepackage[hyperref]{emnlp2020}
\usepackage{times}
\usepackage{latexsym}
\usepackage{amsfonts}

\usepackage[normalem]{ulem}
\useunder{\uline}{\ul}{}
\usepackage{amssymb}
\usepackage{amsfonts}
\usepackage{graphicx}
\usepackage{framed}
\usepackage{colortbl}
\usepackage{color}
\usepackage{multicol}
\usepackage{multirow}
\usepackage{makecell}
\usepackage{amsmath}
\usepackage{booktabs}
\usepackage{array}
\usepackage{arydshln}
\usepackage{paralist}
\usepackage{algorithm}
\usepackage[noend]{algpseudocode}
\allowdisplaybreaks[4]

\definecolor{gray_low}{gray}{.9}
\definecolor{gray_mid}{gray}{.8}
\definecolor{gray_high}{gray}{.7}

\usepackage{arydshln}

\usepackage{microtype}

\aclfinalcopy 

\usepackage{CJKutf8}

\title{Feature Adaptation of Pre-Trained Language Models \\ across Languages and Domains with Robust Self-Training}

\author{Hai Ye\textsuperscript{\rm 1} \ \ \
Qingyu Tan\thanks{~~Qingyu Tan is under the Joint PhD Program between Alibaba and National University of Singapore.}\textsuperscript{\rm ~~1,2} \ \ \
Ruidan He\textsuperscript{\rm 2} \ \ \
\textbf{Juntao Li}\textsuperscript{\rm 3}\\
\textbf{Hwee Tou Ng}\textsuperscript{\rm 1} \ \ \  
\textbf{Lidong Bing}\textsuperscript{\rm 2}
\\
\textsuperscript{\rm 1}Department of Computer Science, National University of Singapore\\
\textsuperscript{\rm 2}DAMO Academy, Alibaba Group\\
\textsuperscript{\rm 3}School of Computer Science and Technology, Soochow University\\
\texttt{\{yeh,nght\}}\texttt{@comp.nus.edu.sg} \\ \texttt{\{qingyu.tan,ruidan.he,l.bing\}}\texttt{@alibaba-inc.com} \\
\texttt{ljt@suda.edu.cn}
\\
}

\date{}
\begin{document}
\maketitle
\begin{abstract}

Adapting pre-trained language models (PrLMs)~(e.g., BERT) to new domains has gained much attention recently. Instead of fine-tuning PrLMs as done in most previous work, we investigate how to adapt the features of PrLMs to new domains without fine-tuning. We explore unsupervised domain adaptation (UDA) in this paper. With the features from PrLMs, we adapt the models trained with labeled data from the source domain to the unlabeled target domain. Self-training is widely used for UDA, and it predicts pseudo labels on the target domain data for training. However, the predicted pseudo labels inevitably include noise, which will negatively affect training a robust model. To improve the robustness of self-training, in this paper we present class-aware feature self-distillation (CFd) to learn discriminative features from PrLMs, in which PrLM features are self-distilled into a feature adaptation module and the features from the same class are more tightly clustered. We further extend CFd to a cross-language setting, in which language discrepancy is studied. 
Experiments on two monolingual and multilingual Amazon review datasets show that CFd can consistently improve the performance of self-training in cross-domain and cross-language settings~\footnote{The source code of the paper is available at \url{https://github.com/oceanypt/CFd}.}.

\end{abstract}

\section{Introduction}

Pre-trained language models~(PrLMs) such as BERT~\cite{bert} and its variants~\cite{robert,xlnet} have shown significant success for various downstream NLP tasks. However, these deep neural networks are sensitive to different cross-domain distributions~\cite{datashift} and their effectiveness will be much weakened in such a scenario. How to adapt PrLMs to new domains is important. 
Unlike the most recent work that fine-tunes PrLMs on the unlabeled data from the new domains~\cite{DBLP:conf/emnlp/HanE19,DBLP:journals/corr/abs-2004-10964}, we are interested in how to adapt the PrLM features without fine-tuning. To investigate this, we specifically study unsupervised domain adaptation~(UDA) of PrLMs, in which we adapt the models trained with source labeled data to the unlabeled target domain based on the features from PrLMs. 

Self-training has been proven to be effective in UDA~\cite{tri-net}, which uses the model trained with source labeled data to predict pseudo labels on the unlabeled target set for model training. Unlike the methods of adversarial learning~\cite{adv,chen2018adversarial} and Maximum Mean Discrepancy~(MMD)~\cite{DBLP:journals/jmlr/GrettonBRSS12} that learn domain-invariant features for domain alignment, self-training aims to learn discriminative features over the target domain, since simply matching domain distributions cannot make accurate predictions on the target after adaptation~\cite{DBLP:conf/iccv/LeeKKJ19,tri-net}. To learn discriminative features for the target, self-training needs to retain a model's high-confidence predictions on the target domain which are considered correct for training. Methods like ensemble learning \cite{DBLP:conf/iccv/ZouYLKW19,DBLP:conf/iclr/GeCL20,tri-net} which adopt multiple models to jointly make decisions on pseudo-label selections have been introduced to achieve this goal. Though these methods can substantially reduce wrong predictions on the target, there will still be noisy labels in the pseudo-label set, with negative effects on training a robust model, since deep neural networks with their high capacity can easily fit to corrupted labels~\cite{DBLP:conf/icml/ArpitJBKBKMFCBL17}. 

In our work, to improve the robustness of self-training, we propose to jointly learn discriminative features from the PrLM on the target domain to alleviate the negative effects caused by noisy labels. We introduce class-aware feature self-distillation~(CFd) to achieve this goal~($\S$\ref{sec:CFd}). The features from PrLMs have been proven to be highly discriminative for downstream tasks, so we propose to distill this kind of features to a feature adaptation module~(FAM) to make FAM capable of extracting discriminative features~($\S$\ref{sec:fd}). 
Inspired by recent work on representation learning~\cite{DBLP:journals/corr/abs-1807-03748,DBLP:conf/iclr/HjelmFLGBTB19}, we introduce mutual information~(MI) maximization for feature self-distillation~(Fd). We maximize the MI between the features from the PrLM and the FAM to make the two kinds of features more dependent. 
Since Fd can only distill features from the PrLM, it ignores the cluster information of data points which can also improve feature discriminativeness~\cite{DBLP:conf/aistats/ChapelleZ05,DBLP:conf/iccv/LeeKKJ19}.   
Hence, for the features output by FAM, if the corresponding data points belong to the same class, we further minimize their feature distance to make the cluster more cohesive, so that different classes will be more separable. 
To retain high-confidence predictions, we  
re-rank the predicted candidates and balance the numbers of samples in different classes~($\S$\ref{sec:self-train}). 

We use XLM-R~\cite{conneau2019unsupervised} as the PrLM which is trained on over 100 languages. 
We also extend our method to cross-language, as well as cross-language and cross-domain settings using XLM-R, since it has already mapped different languages into a common feature space. 
We experiment with two monolingual and multilingual Amazon review datasets for sentiment classification: MonoAmazon for cross-domain and MultiAmazon for cross-language experiments. We demonstrate that self-training can be consistently improved by CFd in all settings~($\S$\ref{sec:main-results}). Further empirical results indicate that the improvements come from learning lower errors of ideal joint hypothesis~($\S$\ref{sec:analysis},\ref{sec:further-analysis}).

\section{Related Work}

\noindent{\textbf{Adaptation of PrLMs.}} Recently, significant improvements on multiple NLP tasks have been enabled by pre-trained language models~(PrLMs)~\cite{bert,xlnet,robert,DBLP:conf/acl/RuderH18,DBLP:conf/naacl/PetersNIGCLZ18}. To enhance their performance on new domains, much work has been done to adapt PrLMs. Two main adaptation settings have been studied. The first is the same as what we study in this work: the PrLM provides the features based on which domain adaptation is conducted
~\cite{DBLP:conf/emnlp/HanE19,DBLP:journals/corr/abs-1911-06137,DBLP:conf/acl/LogeswaranCLTDL19,DBLP:conf/acl-deeplo/MaXWNX19,li2020unsupervised}. In the second setting, the corpus for pre-training a language model has large domain discrepancy with the target domain, so in this scenario, we need the target unlabeled data to fine-tune the PrLM after which we train a task-specific model~\cite{DBLP:journals/corr/abs-2004-10964}. 
For example, \citet{DBLP:journals/bioinformatics/LeeYKKKSK20} and \citet{DBLP:journals/corr/abs-1904-03323} transfer PrLMs into biomedical and clinical domains. Instead of fine-tuning PrLMs with unlabeled data from the new domain as in most previous work~\cite{DBLP:journals/corr/abs-1908-11860,DBLP:conf/emnlp/HanE19,DBLP:journals/corr/abs-2004-10964}, we are interested in the feature-based approach~\cite{bert,peters2019tune} to adapt PrLMs, which does not fine-tune PrLMs. The feature-based approach is much faster, easier, and more memory-efficient for training than the fine-tuning-based method, since it does not have to update the parameters of the PrLMs which are usually massive especially the newly released GPT-3~\cite{brown2020language}. 

\noindent{\textbf{Domain Adaptation.}} To perform domain adaptation, previous work mainly focuses on how to minimize the domain discrepancy and how to learn discriminative features on the target domain~\cite{DBLP:journals/ml/Ben-DavidBCKPV10}. Kernelized methods, e.g., MMD~\cite{DBLP:journals/jmlr/GrettonBRSS12,DBLP:conf/icml/LongC0J15}, and adversarial learning~\cite{adv,chen2018adversarial} are commonly used to learn domain-invariant features. To learn discriminative features for DA, self-training is widely explored~\cite{tri-net,DBLP:conf/iclr/GeCL20,DBLP:conf/iccv/ZouYLKW19,DBLP:conf/eccv/ZouYKW18,DBLP:conf/emnlp/HeLND18}. To retain high-confidence predictions for self-training, ensemble methods like tri-training \cite{tri-net}, mutual learning \cite{DBLP:conf/iclr/GeCL20} and dual information maximization~\cite{ye2019jointly} have been introduced. However, the pseudo-label set will still have noisy labels which will negatively affect model training~\cite{DBLP:conf/icml/ArpitJBKBKMFCBL17,DBLP:conf/iclr/ZhangBHRV17}. Other methods on learning discriminative features include feature reconstruction~\cite{DBLP:conf/eccv/GhifaryKZBL16}, semi-supervised learning~\cite{DBLP:conf/iclr/LaineA17}, and virtual adversarial training~\cite{DBLP:conf/iccv/LeeKKJ19}. Based on cluster assumption~\cite{DBLP:conf/aistats/ChapelleZ05} and the relationship between decision boundary and feature representations, \citet{DBLP:conf/iccv/LeeKKJ19} explore class information to learn discriminative features. Class information is also studied in distant supervision learning for relation extraction~\cite{DBLP:conf/acl/YeCLL17}. 
In NLP, early work explores domain-invariant and domain-specific words to reduce domain discrepancy~\cite{DBLP:conf/acl/BlitzerDP07,DBLP:conf/www/PanNSYC10,DBLP:conf/acl/HeLA11}.

\section{Preliminary}

In this section, we introduce the problem definition and the model architecture based on which we build our domain adaptation algorithm presented in the next section.

\subsection{Unsupervised Domain Adaptation}

In order to improve the feature adaptability of pre-trained transformers cross domains, we study unsupervised domain adaptation of pre-trained language models where we train models with labeled data and unlabeled data from the source and target domain respectively. We use the features from PrLMs to perform domain adaptation. Labeled data from the source domain are defined as $S = \{X_s, Y_s\}$, in which every sample $\mathbf{x}_s \in X_s$ has a label $\mathbf{y}_s \in Y_s$. The unlabeled data from the target domain are $T = \{X_t\}$. 
In this work, we comprehensively study domain adaptation in cross-domain and cross-language settings, based on the features from the multi-lingual PrLM where we adopt XLM-R~\cite{conneau2019unsupervised} for evaluation. By using XLM-R, different languages can be mapped into a common feature space. In this work, we evaluate our method on the task of sentiment classification using two datasets. 

\begin{figure}[t]
\setlength{\abovecaptionskip}{-0.1cm}
\setlength{\belowcaptionskip}{-0.2cm}
\begin{center}
\includegraphics[width=\columnwidth]{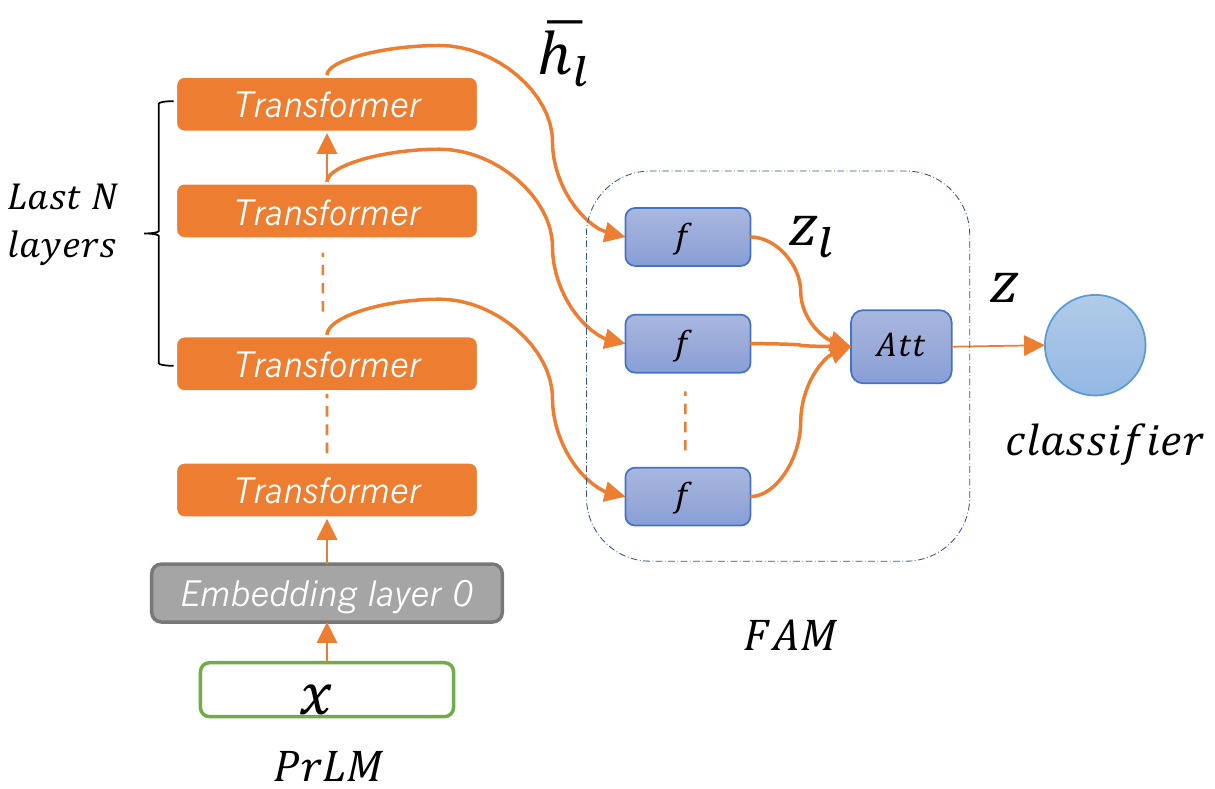}
\end{center}
\caption{Illustration of our model architecture which includes a pre-trained language model, a feature adaptation module, and a classifier.}
\label{fig:model}
\end{figure}

\subsection{Model Architecture}
As presented in Figure~\ref{fig:model}, our model consists of a pre-trained language model~(PrLM), a feature adaptation module~(FAM), and a classifier. 

\subsubsection{Pre-trained Language Model}
Following BERT~\cite{bert}, most PrLMs consist of an embedding layer and several transformer layers. Suppose a PrLM has $L+1$ layers, layer $0$ is the embedding layer, and layer $L$ is the last layer. 
Given an input sentence $\mathbf{x} = [w_1, w_2, \cdots, w_{|\mathbf{x}|}]$, the embedding layer of the PrLM will encode $\mathbf{x}$ as: 
\begin{equation}
\setlength\abovedisplayskip{4pt}\setlength\belowdisplayskip{4pt}
    \mathbf{h}_0 = \text{Embedding}(\mathbf{x})
\end{equation}
where $\mathbf{h}_0 = [\mathbf{h}_0^{1}, \mathbf{h}_0^{2}, \cdots, \mathbf{h}_0^{|\mathbf{x}|}]$.
After obtaining the embeddings of the input sentence, we compute the features of the sentence from the transformer blocks of PrLM. In layer $l$, we compute the transformer feature as:
\begin{equation}
\setlength\abovedisplayskip{4pt}\setlength\belowdisplayskip{4pt}
    \mathbf{h}_l = \text{Transformer}_{l}(\mathbf{h}_{l-1})
\end{equation}
where $\mathbf{h}_l$ $=$ $[\mathbf{h}_l^{1}, \mathbf{h}_l^{2}, \cdots, \mathbf{h}_{l}^{|\mathbf{x}|}]$ and $l$ $\in$ $\{1, 2,$ $\cdots, L\}$. 
Using all the $|\mathbf{x}|$ features will incur much memory space. After experiments, we take the average of $\mathbf{h}_l$ as:
\begin{equation}
\setlength\abovedisplayskip{4pt}\setlength\belowdisplayskip{4pt}
    \bar{\mathbf{h}}_l = \frac{1}{|\mathbf{x}|} \sum_{i=1}^{|\mathbf{x}|} \mathbf{h}_l^i
\end{equation}
and $\bar{\mathbf{h}}_l$ will be fed into the FAM. 

\subsubsection{Feature Adaptation Module}\label{sec:FAM}
To transfer the knowledge from the source to the target domain, the features from PrLMs should be more transferable. 
Previous work points out that the PrLM features from the intermediate layers are more transferable than the upper-layer features, and the upper-layer features are more discriminative for classification~\cite{hao-etal-2019-visualizing,DBLP:conf/naacl/PetersNIGCLZ18,DBLP:conf/naacl/Liu0BPS19}. By making a trade-off between speed and model performance, we combine the last $N$-layer features from the PrLM for domain adaptation, which is called the multi-layer representation of the PrLM. 

Our FAM consists of a feed-forward neural network~(followed by a $\tanh$ activation function) and an attention mechanism. We map $\bar{\mathbf{h}}_l$ from layer $l$ into $\mathbf{z}_l$ with the feed-forward neural network:
\begin{equation}
\setlength\abovedisplayskip{4pt}\setlength\belowdisplayskip{4pt}
\begin{split}
    \mathbf{z}_l = f(\bar{\mathbf{h}}_l)
\end{split}
\end{equation}

\noindent{\textbf{Multi-layer Representation.}} Since feature effectiveness differs from layer to layer, we use an attention mechanism~\cite{DBLP:conf/emnlp/LuongPM15} to learn to weight the features from the last $N$ layers. We get the multi-layer representation $\mathbf{z}$ of the PrLM as:
\begin{equation}
\setlength\abovedisplayskip{4pt}\setlength\belowdisplayskip{4pt}
    \begin{split}
            \mathbf{z} &= E(\mathbf{x}; \theta) = \sum_{i=L-N+1}^L \alpha_i \mathbf{z}_i \\
            \alpha_i &= \frac{e^{\tanh(\mathbf{W}_{att}\mathbf{z}_i)}}{\sum_{j=L-N+1}^L  e^{\tanh(\mathbf{W}_{att}\mathbf{z}_j)}}
    \end{split}
\end{equation}
in which $\mathbf{W}_{att}$ is a matrix of trainable parameters. 
Inspired by \citet{DBLP:conf/nips/BerthelotCGPOR19}, we want the model to focus more on the higher-weighted layers, so we further calculate the attention weight as:
\begin{equation}
\setlength\abovedisplayskip{4pt}\setlength\belowdisplayskip{4pt}
    \alpha_i = \frac{\alpha_i^{1/\tau}}{\sum_{j=L-N+1}^{L}\alpha_j^{1/\tau}}
    \label{eq:tau}
\end{equation}
where $\theta$ is a set of learnable parameters that includes the parameters from the feed-forward neural network and the attention mechanism.

\subsubsection{Classifier}
After obtaining the multi-layer representation $\mathbf{z}$, we train a classifier with the source domain labeled set $S$. We define the loss function for the task-specific classifier as:
\begin{equation}
\setlength\abovedisplayskip{4pt}\setlength\belowdisplayskip{4pt}
\begin{split}
     \mathcal{L}_{pred}^S = \frac{1}{|S|}\sum_{\langle \mathbf{x}, \mathbf{y} \rangle \in S} l \big (g(E(\mathbf{x};\theta);\phi), \mathbf{y} \big ) 
\end{split}
\label{eq:source}
\end{equation}
where $g$ is a classifier that takes in the features out of $E$, and $g$ is parameterized by $\phi$. $l$ is the loss function which is cross-entropy loss in our work. 

\section{Class-aware Feature Self-distillation for Domain Adaptation}
In this section, we introduce our method for domain adaptation. Our domain adaptation loss function takes the form of:
\begin{equation}
\setlength\abovedisplayskip{4pt}\setlength\belowdisplayskip{4pt}
    \mathcal{L} = \mathcal{L}_{pred}^S + \mathcal{L}_{pred}^{T'} + \mathcal{L}_{CFd}
\end{equation}
in which $\mathcal{L}_{pred}^S$ is for learning a task-specific classifier with the source labeled set $S$~(Eq.~\ref{eq:source}), $\mathcal{L}_{pred}^{T'}$ is the self-training loss trained with the pseudo-label set $T'$~($\S$\ref{sec:self-train}), and $\mathcal{L}_{CFd}$ is to enhance the robustness of self-training by learning discriminative features from the PrLM~($\S$\ref{sec:CFd}), which is the main algorithm for domain adaptation in this work. 

\subsection{Self-training for Adaptation}\label{sec:self-train}
We build our adaptation model based on self-training, which predicts pseudo labels on unlabeled target data. The predicted pseudo labels will be used for model training. In the training process, we predict pseudo labels on all the target samples in $T$. To retain high-confidence predictions from $T$, 
we introduce a simple but effective method called \emph{rank-diversify} to build the pseudo-label set $T'$ , which is a subset of $T$:

\emph{\textbf{Rank.}} We calculate the entropy loss for every sample in $T$, specifically:
\begin{equation}
\setlength\abovedisplayskip{4pt}\setlength\belowdisplayskip{4pt}
\begin{split}
    g(\mathbf{z}) &= Softmax \big (g(\mathbf{z}) \big ) \\
    \mathcal{L}_{e}(\mathbf{z}) &= -\sum g(\mathbf{z})^{T}\log g(\mathbf{z})
\end{split}
\end{equation}
in which $\mathbf{z}$ is the multi-layer feature and $g$ is the classifier in~Eq.~\ref{eq:source}. A lower entropy loss indicates a higher confidence of the model for the pseudo label. Then we use the entropy loss to re-rank $T$. However, after re-ranking, some classes may have too many samples in the top $K$ candidates, which will bias model training, so we also need to diversify the pseudo labels in the top $K$ list.

\emph{\textbf{Diversify.}} We classify the samples into different classes with pseudo labels, and re-rank them with entropy loss in ascending order in every class. Samples are selected following the order from every class in turn until $K$ samples are selected.  

With the retained pseudo-label set $T'$, we have the loss function for training as:
\begin{equation}
\setlength\abovedisplayskip{4pt}\setlength\belowdisplayskip{4pt}
    \mathcal{L}_{pred}^{T'} = \alpha \frac{1}{|T'|}\sum_{\langle \mathbf{x}, \mathbf{y} \rangle \in T'} l \big (g(E(\mathbf{x};\theta);\phi),\mathbf{y} \big )
\end{equation}
in which $\alpha$ is a hyper-parameter which will increase gradually in the training process. 

\begin{figure}[t]
\setlength{\abovecaptionskip}{-0.1cm}
\setlength{\belowcaptionskip}{-0.2cm}
\begin{center}
\includegraphics[width=5.5cm]{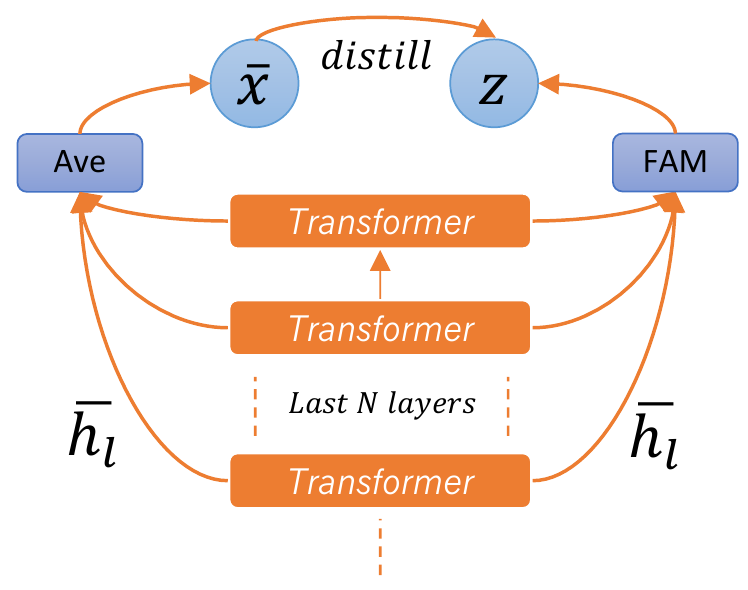}
\end{center}
\caption{Illustration of feature self-distillation. We take the sum of the last $N$-layer features for distillation.}
\label{fig:mi}
\end{figure}

\subsection{Robust Self-training by Discriminative Feature Learning}\label{sec:CFd}
To alleviate the negative effects caused by the noisy labels in the pseudo-label set $T'$, we propose to learn discriminative features from the PrLM.

\subsubsection{Feature Self-distillation}\label{sec:fd}

To maintain the discriminative power of PrLM features, we propose to self-distill the PrLM features into the newly added feature adaptation module~(FAM). Similar to traditional knowledge distillation~\cite{DBLP:journals/corr/HintonVD15}, feature distillation in our work is to make the FAM~(\emph{student}) also capable of generating discriminative features for adaptation as the PrLM~(\emph{teacher}) does. 
Since the source domain already has the labeled data, there is no need for self-distillation on the source domain, and we apply feature self-distillation~(Fd) to the target domain. Inspired by recent work on representation learning~\cite{DBLP:journals/corr/abs-1807-03748,DBLP:conf/iclr/HjelmFLGBTB19,DBLP:conf/iclr/TianKI20}, we propose to use mutual information~(MI) maximization for Fd.

\noindent{\textbf{MI for Feature Self-distillation.}} MI measures how different two random variables are. Maximizing the MI between them can reduce their difference. By maximizing the MI between the features from PrLM and FAM, we can make the two features more similar. 
We are interested in distilling the PrLM features into the multi-layer representation $\mathbf{z}$. 
We can distill the feature $\bar{\mathbf{h}}_l$ from any layer $l$ into $\mathbf{z}$. 
However, only distilling one-layer feature of the PrLM may neglect the information from other layers, so we use the sum of the last $N$-layer features for distillation\footnote{Based on Eq.\ref{eq:cos}, taking the sum or average of the last $N$-layer features will have the same effect.}:
\begin{equation}
\setlength\abovedisplayskip{4pt}\setlength\belowdisplayskip{4pt}
    \bar{\mathbf{x}} = \sum_{i=L-N+1}^L \bar{\mathbf{h}}_i
\end{equation}
The distillation process is illustrated in Figure~\ref{fig:mi}. Then we maximize the MI $I(\mathbf{z}, \bar{\mathbf{x}})$. We need to find its lower bound for maximization, since it is hard to directly estimate mutual information. 
Following~\citet{DBLP:journals/corr/abs-1807-03748}, we also use Noise Contrastive Estimation~(NCE) to infer the lower bound as: 
\begin{equation}
\setlength\abovedisplayskip{4pt}\setlength\belowdisplayskip{4pt}
    I(\mathbf{z}, \bar{\mathbf{x}}) \ge \mathcal{J}^{feat}_{NCE}(\mathbf{z}, \bar{\mathbf{x}}) 
    \label{lower-bound-feat}
\end{equation}
To estimate the NCE loss, we need a negative sample set in which the PrLM features are randomly sampled for the current $\mathbf{z}$. Given a negative sample set $\bar{X}^{neg}$ $=$ $\{\bar{\mathbf{x}}_i^{neg}\}_{i=1}^{|\bar{X}^{neg}|}$, we estimate $\mathcal{J}^{feat}_{NCE}$ as:
\begin{equation}
\setlength\abovedisplayskip{4pt}\setlength\belowdisplayskip{4pt}
    \mathcal{J}_{NCE}^{feat} = f(\mathbf{z}, \bar{\mathbf{x}}) - \frac{1}{|\bar{X}^{neg}|} \sum_{\bar{\mathbf{x}}_i^{neg} \in \bar{X}^{neg}} f(\mathbf{z}, \bar{\mathbf{x}}_i^{neg})
     \label{nce}
\end{equation}
$f(\mathbf{z}, \bar{\mathbf{x}}^*)$ 
is the similarity function, defined as: 
\begin{equation}
\setlength\abovedisplayskip{4pt}\setlength\belowdisplayskip{4pt}
    f(\mathbf{z}, \bar{\mathbf{x}}^*) = { \cos \big (\text{inf}(\mathbf{z}), \bar{\mathbf{x}}^* \big )}\label{eq:cos}
\end{equation}
where $\bar{\mathbf{x}}^* \in \{\bar{\mathbf{x}}\} \cup \bar{X}^{neg}$; $\text{inf}(\cdot)$ is a trainable feed-forward neural network followed by the $\tanh$ activation, which is to resize the dimension of $\mathbf{z}$ to be equal to $\bar{\mathbf{x}}^*$. To obtain the negative sample set, we select one negative $\bar{\mathbf{x}}$ by randomly
shuffling the batch of features which the negative $\bar{\mathbf{x}}$ is in, and this process is repeated $|\bar{X}^{neg}|$ times. 

\subsubsection{Class Information}\label{sec:c}

Feature distillation can only maintain the discriminative power of PrLM features but ignores the class information present in class labels. To explore the class information, when performing feature self-distillation, we further introduce an intra-class loss to minimize the feature distance under the same class. By giving the pseudo-label set $T'$ and the source labeled set $S$, we group the multi-layer features out of the FAM into different classes. For every class $c$, we calculate the center feature as $\mathbf{z}_c$.  
We define the intra-class loss as follows:
\begin{equation}
\setlength\abovedisplayskip{4pt}\setlength\belowdisplayskip{4pt}
    \mathcal{L}_{intra\_class} = \sum_{c \in C} \sum_{\mathbf{z}_i \in S_c \cup T'_c}  \lVert \mathbf{z}_i - \mathbf{z}_c \rVert_2
\end{equation}
where $C$ is the set of classes. The center feature $\mathbf{z}_c$ for class $c \in C$ is calculated as:
\begin{equation}
\setlength\abovedisplayskip{4pt}\setlength\belowdisplayskip{4pt}
    \mathbf{z}_c = \frac{1}{|S_c \cup T'_c|}  \sum_{\mathbf{z}_j \in S_c \cup T'_c}   \mathbf{z}_j
\end{equation}
Before training for an epoch, the center features will be calculated and fixed during training. After one epoch of training, the center features will be updated. After the above analysis, our final CFd loss becomes:
\begin{equation}
\setlength\abovedisplayskip{4pt}\setlength\belowdisplayskip{2pt}
\begin{split}
        \mathcal{L}_{CFd} = \mathcal{L}_{Fd}^{T} + \mathcal{L}_{C}^{S,T'} &=  -\sum\limits_{\mathbf{x} \in T} \mathcal{J}_{NCE}^{feat} \big (E(\mathbf{x};\theta), \bar{\mathbf{x}} \big ) \\  &+ \lambda \sum\limits_{\langle \mathbf{x}, \mathbf{y} \rangle \in S \cup T'} \mathcal{L}_{intra\_class}
\end{split}
\end{equation}
where $\lambda$ is a hyper-parameter which controls the contribution of $\mathcal{L}_{C}^{S,T'}$.

\begin{table*}[t]
\setlength{\abovecaptionskip}{0.2cm}
\setlength{\belowcaptionskip}{-0.2cm}
\centering
\resizebox{\textwidth}{!}{
\setlength{\tabcolsep}{0.8mm}{\begin{tabular}{l lllllllllllll}
\toprule
\textbf{MonoAmazon} & E$\to$BK            & BT$\to$BK & M$\to$BK & BK$\to$E & BT$\to$E & M$\to$E & BK$\to$BT & E$\to$BT & M$\to$BT & BK$\to$M & E$\to$M & BT$\to$M & Ave.                                  \\ \hline 
{\color[HTML]{000000} DAS}        & {\color[HTML]{000000} 67.12}                        & {\color[HTML]{000000} 66.53}                         & {\color[HTML]{000000} 70.31}                        & {\color[HTML]{000000} 58.73}                        & {\color[HTML]{000000} 66.14}                        & {\color[HTML]{000000} 55.78}                       & {\color[HTML]{000000} 51.30}                          & {\color[HTML]{000000} 60.76}                        & {\color[HTML]{000000} 50.66}                        & {\color[HTML]{000000} 55.98}                        & {\color[HTML]{000000} 59.06}                       & {\color[HTML]{000000} 60.50}                         & {\color[HTML]{000000} 60.24}      \\ \cdashline{1-14}

xlmr-tuning & 70.03$_{0.2}$ & 69.94$_{0.9}$ & 70.71$_{0.8}$ & 61.27$_{0.5}$ & 68.49$_{0.4}$ & 63.52$_{1.0}$ & 66.27$_{1.3}$ & 69.81$_{1.2}$ & 68.32$_{0.6}$ & 61.69$_{2.5}$ & 59.22$_{1.1}$ & 61.75$_{1.9}$ & 65.92 \\

xlmr-1     & 64.70           & 64.26               & 68.64              & 53.21              & 66.39              & 55.67             & 57.88               & 70.10              & 55.20               & 61.05              & 63.92             & 65.60              & 63.52                                 \\ 
xlmr-10    & 70.58$_{0.3}$           & 69.96$_{0.6}$               & 71.10$_{0.5}$              & 59.80$_{0.3}$              & 70.88$_{0.3}$              & 64.64$_{0.7}$             & 63.93$_{0.9}$               & 72.48$_{0.5}$              & 65.06$_{0.9}$              & 65.79$_{0.4}$              & 67.78$_{0.4}$             & 63.49$_{1.0}$              &   67.12                               \\ 
 
KL         & 70.91$_{0.7}$           & 71.12$_{0.3}$               & 72.10$_{0.3}$              & 65.61$_{0.1}$              & 70.30$_{0.5}$              & 66.85$_{0.4}$             & 67.69$_{0.7}$               & 72.68$_{0.2}$              & 70.36$_{0.3}$              & 67.66$_{0.7}$              & 66.46$_{1.1}$             & 68.56$_{1.1}$              & {\color[HTML]{000000} 69.19}          \\ 
MMD        & 71.91$_{0.7}$           & 73.58$_{0.6}$               & 70.48$_{0.8}$              & 69.37$_{0.6}$              & 71.27$_{0.5}$              & 65.92$_{0.9}$             & \textbf{71.71}$_{0.5}$      & 72.81$_{0.5}$              & 69.30$_{0.5}$              & 69.24$_{0.5}$              & 65.87$_{1.0}$             & 69.14$_{1.0}$              & {\color[HTML]{000000} 70.05}          \\ 
Adv        & 71.28$_{0.5}$           & 69.53$_{1.0}$               & 72.39$_{0.2}$              & 61.20$_{0.6}$              & 69.98$_{0.4}$              & 66.47$_{0.2}$             & 63.91$_{1.3}$               & 72.84$_{0.3}$              & 70.47$_{0.1}$              & 66.53$_{0.7}$              & 67.65$_{0.4}$             & 64.47$_{1.6}$              & {\color[HTML]{000000} 68.06}          \\ \hline
                               
p          & 70.90$_{0.4}$           & 71.38$_{0.8}$               & 72.18$_{0.9}$              & 64.00$_{1.2}$              & 70.41$_{0.5}$              & 67.01$_{0.3}$             & 67.48$_{0.4}$               & 71.67$_{0.5}$              & \textbf{70.71}$_{0.3}$     & 67.16$_{0.6}$              & 67.92$_{1.1}$             & 69.77$_{0.2}$              & 69.21                                 \\

\textbf{p+CFd}      & \textbf{75.25}$_{0.5}$ & \textbf{74.70}$_{0.5}$     & \textbf{75.08}$_{0.6}$    & \textbf{70.19}$_{0.2}$    & \textbf{72.00}$_{0.3}$    & \textbf{68.96}$_{0.3}$   & 71.63$_{0.4}$               & \textbf{73.73}$_{0.5}$    & 70.05$_{0.4}$              & \textbf{70.86}$_{0.3}$    & \textbf{69.80}$_{0.7}$   & \textbf{70.46}$_{0.4}$    & {\color[HTML]{000000} \textbf{71.89}} \\ 

\bottomrule

\end{tabular}}
}
\caption{The cross-domain classification accuracy~(\%) results on MonoAmazon. Models are evaluated by 5 random runs except xlmr-tuning which is run for 3 times. We report the mean and standard deviation results. Best task performance is boldfaced. 
Results of DAS are taken from \citet{DBLP:conf/emnlp/HeLND18}.}
\label{Tab:cross-domain}
\end{table*}

\subsection{Analysis}\label{sec:analysis}
We provide a theoretical understanding for why CFd can enhance self-training based on the domain adaptation theory from \citet{DBLP:journals/ml/Ben-DavidBCKPV10}.

\noindent{\textbf{Theorem 1.}}~\cite{DBLP:journals/ml/Ben-DavidBCKPV10} \emph{Let $\mathcal{H}$ be the hypothesis space. With the generalization error $\delta_s$ and $\delta_t$ of a classifier $G \in \mathcal{H}$ on the source $S$ and target $T$, we have:}
\begin{equation}
\setlength\abovedisplayskip{4pt}\setlength\belowdisplayskip{4pt}
    \delta_t(G) \le \delta_s(G) + d_{\mathcal{H}\Delta\mathcal{H}}(S,T) + \epsilon
    \label{eq:da}
\end{equation}
in which $d_{\mathcal{H}\Delta\mathcal{H}}$ measures the domain discrepancy and is defined as:
\begin{equation}
\setlength\abovedisplayskip{4pt}\setlength\belowdisplayskip{4pt}
\begin{split}
    d_{\mathcal{H}\Delta\mathcal{H}}(S,T) = \sup\limits_{h, h' \in \mathcal{H}}\big |\mathbb{E}_{\mathbf{x} \in S}[h(\mathbf{x}) \ne h'(\mathbf{x})] \\ - \mathbb{E}_{\mathbf{{x}} \in T}[h(\mathbf{x}) \ne h'(\mathbf{x})] \big |
\end{split}
\end{equation}
and $\epsilon$ is the error of the ideal joint hypothesis which is defined as:
\begin{equation}
\setlength\abovedisplayskip{2pt}\setlength\belowdisplayskip{2pt}
    \epsilon = \delta_s(h^*) + \delta_t(h^*)
    \label{eq:joint_error}
\end{equation}
where $h^* = \arg \min_{h \in \mathcal{H}}\delta_s(h) + \delta_t(h)$.

From Ineq.~\ref{eq:da}, the performance of domain adaptation is bounded by the generalization error on the source domain, domain discrepancy, and the error of the ideal joint hypothesis~(joint error). Self-training aims to learn a low joint error by learning discriminative features on the target domain, so that the adaptation performance can be improved~\cite{tri-net}. Our proposed CFd enhances the robustness of self-training by self-distilling the PrLM features and exploring the class information. In this way, the joint error can be further reduced compared to self-training~(Fig.~\ref{fig:joint_error}). Besides, by optimizing the intra-class loss, $d_{\mathcal{H}\Delta\mathcal{H}}$ in Ineq.~\ref{eq:da} can be reduced since under the same class, the feature distance of samples from both the source and target domain is minimized~(Fig.~\ref{fig:a_distance}).

\begin{table}[]
\setlength{\abovecaptionskip}{0.2cm}
\setlength{\belowcaptionskip}{-0.2cm}
\centering
\setlength{\tabcolsep}{1mm}{\resizebox{\columnwidth}{!}{
\begin{tabular}{lccccc}
\toprule

  \sc Data          & train~(\emph{S}) & valid~(\emph{S}) & test~(\emph{T}) & unlabeled~(\emph{T}) & $|C|$ \\ \hline
MonoAmazon  & 5,000     & 1,000     & 6,000   & 6,000  & 3      \\ 
MultiAmazon & 2,000     & 2,000     & 2,000   & 8,000   & 2 \\ \toprule
\end{tabular}}
}
\caption{The data splits for the experiments. $|C|$ is the number of classes. ($\cdot$) denotes the domain which the data comes from.}
\label{Tab:dataset}
\end{table}

\begin{table*}[t]
\setlength{\abovecaptionskip}{0.2cm}
\setlength{\belowcaptionskip}{-0.2cm}
\centering
\resizebox{\textwidth}{!}{
\begin{tabular}{l cccc cccc cccc}
\toprule
                                            & \multicolumn{3}{c}{ German}                                                                                             &                                       & \multicolumn{3}{c}{ French}                                                                                               &                                       & \multicolumn{3}{c}{ Japanese}                                                                                           &                                       \\ \cline{2-4} \cline{6-8} \cline{10-12}
\multirow{-2}{*}{\textbf{MultiAmazon}} &  Book                                  & Dvd                                    &  Music                                 & \multirow{-2}{*}{ Ave.}                & Book                                   & Dvd                                    & Music                                  & \multirow{-2}{*}{ Ave.}                &  Book                                  & Dvd                                   & Music                                  & \multirow{-2}{*}{ Ave.}                \\ 
\hline
\multicolumn{12}{l}{\textit{\textbf{Cross-language}}}\\

xlmr-tuning & 91.03$_{0.3}$ &	88.02$_{0.6}$ &	90.13$_{0.2}$ &	89.73&	92.12$_{0.5}$&	91.17$_{0.3}$&	89.58$_{0.8}$&	90.96&	87.52$_{0.5}$&	87.12$_{0.4}$&	88.52$_{0.7}$&	87.72 \\

xlmr-1                                      & 73.69                                 & 69.86                                  & 87.34                                 & 76.96                                 & 91.26                                  & 91.13                                  & 88.37                                  & 90.25                                 & 70.96                                 & 71.20                                 & 87.07                                  & 76.41                                 \\ 
xlmr-10                                     & 93.15$_{0.8}$                                 & 89.59$_{1.2}$                                  & 92.26$_{0.6}$                                 & {\color[HTML]{000000} 91.67}          & 93.79$_{0.4}$                                  & 93.28$_{0.4}$                                  & 92.23$_{0.6}$                                  & {\color[HTML]{000000} 93.10}          & {\color[HTML]{000000} 87.13}$_{1.1}$          & {\color[HTML]{000000} 88.63}$_{0.1}$          & {\color[HTML]{000000} 88.05}$_{0.5}$           & {\color[HTML]{000000} 87.94}          \\ 
KL                                          & {\color[HTML]{000000} \textbf{93.99}}$_{0.4}$ & {\color[HTML]{000000} 91.12}$_{0.4}$           & {\color[HTML]{000000} \textbf{93.89}}$_{0.2}$ & {\color[HTML]{000000} 93.00}          & {\color[HTML]{000000} 93.91}$_{0.1}$           & {\color[HTML]{000000} 93.31}$_{0.3}$           & {\color[HTML]{000000} 92.39}$_{0.2}$           & {\color[HTML]{000000} 93.20}          & {\color[HTML]{000000} 88.60}$_{0.1}$          & {\color[HTML]{000000} 88.82}$_{0.2}$          & {\color[HTML]{000000} 88.12}$_{0.2}$           & {\color[HTML]{000000} 88.51}          \\ 
MMD                                         & {\color[HTML]{000000} 93.97}$_{0.1}$          & {\color[HTML]{000000} 90.77}$_{0.8}$           & {\color[HTML]{000000} 93.53}$_{0.4}$          & {\color[HTML]{000000} 92.76}          & {\color[HTML]{000000} 93.48}$_{0.2}$           & {\color[HTML]{000000} 93.21}$_{0.2}$           & {\color[HTML]{000000} 92.67}$_{0.2}$           & {\color[HTML]{000000} 93.12}          & {\color[HTML]{000000} 89.17}$_{0.1}$          & {\color[HTML]{000000} \textbf{89.22}}$_{0.1}$ & {\color[HTML]{000000} 88.54}$_{0.4}$           & {\color[HTML]{000000} 88.98}          \\ 
Adv                                         & {\color[HTML]{000000} 93.27}$_{0.4}$          & {\color[HTML]{000000} 89.78}$_{0.6}$           & {\color[HTML]{000000} 92.53}$_{0.6}$          & {\color[HTML]{000000} 91.86}          & {\color[HTML]{000000} 93.70}$_{0.4}$           & {\color[HTML]{000000} 93.03}$_{0.4}$           & {\color[HTML]{000000} 92.28}$_{0.3}$           & {\color[HTML]{000000} 93.00}          & {\color[HTML]{000000} 88.22}$_{0.8}$          & {\color[HTML]{000000} 88.68}$_{0.1}$          & {\color[HTML]{000000} 88.34}$_{0.2}$           & {\color[HTML]{000000} 88.41}          \\ \hline
p                                           & 92.99$_{1.0}$                                 & 89.33$_{0.6}$                                  & 93.82$_{0.3}$                                 & 92.05                                 & 93.81$_{0.1}$                                  & 93.00$_{0.2}$                                  & 92.50$_{0.2}$                                  & 93.10                                 & 88.68$_{0.3}$                                 & 88.86$_{0.1}$                                 & 88.39$_{0.1}$                                  & 88.64                                 \\ 

\textbf{p+CFd}                              & {\color[HTML]{000000} 93.95}$_{0.2}$          & {\color[HTML]{000000} \textbf{91.69}$_{0.3}$} & {\color[HTML]{000000} \textbf{93.89}}$_{0.2}$ & {\color[HTML]{000000} \textbf{93.18}} & {\color[HTML]{000000} \textbf{94.25}$_{0.2}$} & {\color[HTML]{000000} \textbf{93.79}$_{0.1}$} & {\color[HTML]{000000} \textbf{93.39}$_{0.1}$} & {\color[HTML]{000000} \textbf{93.81}} & {\color[HTML]{000000} \textbf{89.41}$_{0.2}$} & {\color[HTML]{000000} 88.68}$_{0.1}$          & {\color[HTML]{000000} \textbf{89.54}$_{0.3}$} & {\color[HTML]{000000} \textbf{89.21}} \\ 

 \toprule[0.5pt] \toprule[0.5pt] 
\multicolumn{6}{l}{\emph{\textbf{{Cross-language and Cross-domain}}}}  \\

CLDFA &83.95 & 83.14 & 79.02 & 82.04 & 83.37 & 82.56 & 83.31 & 83.08 & 77.36 & 80.52 & 76.46 & 78.11 \\

MAN-MoE & 82.40 & 78.80 & 77.15 & 79.45 & 81.10 & 84.25 & 80.90 & 82.08 & 62.78 & 69.10 & 72.60 & 68.16 \\ \cdashline{1-13}[4pt/2pt]

xlmr-tuning & 90.84 &	88.48	& 89.75 &	89.69 &	90.29 &	90.54 &	89.65 &	90.16 &	85.90 &	86.02 &	87.85 &	86.59  \\

xlmr-1                              & {\color[HTML]{000000} 74.10} & {\color[HTML]{000000} 77.16}
& {\color[HTML]{000000} 66.52} & {\color[HTML]{000000} 72.59} & {\color[HTML]{000000} 87.95} & {\color[HTML]{000000} 88.00} & {\color[HTML]{000000} 88.15} & {\color[HTML]{000000} 88.03} & {\color[HTML]{000000} 76.46} & {\color[HTML]{000000} 75.20} & {\color[HTML]{000000} 65.93} & {\color[HTML]{000000} 72.53} \\ 
xlmr-10                             & {\color[HTML]{000000} 91.00}                     & {\color[HTML]{000000} 85.95}                     & {\color[HTML]{000000} 92.48}                     & {\color[HTML]{000000} 89.81}                     & {\color[HTML]{000000} 90.17}                     & {\color[HTML]{000000} 90.29}                     & {\color[HTML]{000000} 92.66}                     & {\color[HTML]{000000} 91.04}                     & {\color[HTML]{000000} 85.67}                     & {\color[HTML]{000000} 85.69}                     & {\color[HTML]{000000} 87.89}                     & {\color[HTML]{000000} 86.41}                     \\ 
KL                                  & {\color[HTML]{000000} 93.24}                     & {\color[HTML]{000000} 90.39}                     & {\color[HTML]{000000} 93.00}                     & {\color[HTML]{000000} 92.21}                     & {\color[HTML]{000000} 91.98}                     & {\color[HTML]{000000} 92.53}                     & {\color[HTML]{000000} 92.81}                     & {\color[HTML]{000000} 92.44}                     & {\color[HTML]{000000} 86.65}                     & {\color[HTML]{000000} 88.21}                     & {\color[HTML]{000000} 88.61}                     & {\color[HTML]{000000} 87.82}                     \\ 
MMD                                 & {\color[HTML]{000000} 93.44}                     & {\color[HTML]{000000} 90.50}                     & {\color[HTML]{000000} 92.58}                     & {\color[HTML]{000000} 92.17}                     & {\color[HTML]{000000} 92.70}                     & {\color[HTML]{000000} 92.53}                     & {\color[HTML]{000000} 93.07}                     & {\color[HTML]{000000} 92.77}                     & {\color[HTML]{000000} 87.75}                     & {\color[HTML]{000000} 88.25}                     & {\color[HTML]{000000} 88.73}                     & {\color[HTML]{000000} 88.24}                     \\ 
Adv                                 & {\color[HTML]{000000} 92.76}                     & {\color[HTML]{000000} 88.77}                     & {\color[HTML]{000000} 92.80}                     & {\color[HTML]{000000} 91.44}                     & {\color[HTML]{000000} 91.58}                     & {\color[HTML]{000000} 91.70}                     & {\color[HTML]{000000} 92.64}                     & {\color[HTML]{000000} 91.97}                     & {\color[HTML]{000000} 86.88}                     & {\color[HTML]{000000} 88.11}                     & {\color[HTML]{000000} 88.03}                     & {\color[HTML]{000000} 87.67}                     \\ \hline

p                                   & {93.11}                        & {88.43}                        & {92.84}                        & {91.46}                        & {92.09}                        & {92.41}                        & {92.52}                        & {92.34}                        & {87.10}                        & {88.22}                        & {88.57}                        & {87.96}                        \\

\textbf{p+CFd}                      & {\color[HTML]{000000} \textbf{94.29}}           & {\color[HTML]{000000} \textbf{90.73}}            & {\color[HTML]{000000} \textbf{93.62}}           & {\color[HTML]{000000} \textbf{92.88}}            & {\color[HTML]{000000} \textbf{93.10}}            & {\color[HTML]{000000} \textbf{92.81}}            & {\color[HTML]{000000} \textbf{93.62}}            & {\color[HTML]{000000} \textbf{93.18}}            & {\color[HTML]{000000} \textbf{88.93}}           & {\color[HTML]{000000} \textbf{89.00}}            & {\color[HTML]{000000} \textbf{89.41}}           & {\color[HTML]{000000} \textbf{89.11}}   

\\ \bottomrule

\end{tabular}
}
\caption{The classification accuracy~(\%) results on MultiAmazon. 
Models are evaluated by 5 random runs except xlmr-tuning which is run for 3 times. 
Results of CLDFA and MAN-MoE are taken from \citet{DBLP:conf/acl/XuY17} and \citet{DBLP:conf/acl/ChenAHWC19} respectively. More detailed transfer results are included in Appendix~\ref{sec:multiAmazon}.}
\label{Tab:cross-lang}
\end{table*}

\section{Experiments}
\subsection{Datasets}
We use two Amazon review datasets for evaluation. One is monolingual and the other is multilingual.

\noindent{\textbf{MonoAmazon.}} This dataset consists of English reviews from \citet{DBLP:conf/emnlp/HeLND18} and has four domains: Book~(BK), Electronics~(E), Beauty~(BT), and Music~(M). Each domain has 2,000 positive, 2,000 negative, and 2,000 neutral reviews.

\noindent{\textbf{MultiAmazon.}} This is a multilingual review dataset \cite{DBLP:conf/acl/PrettenhoferS10} in English, German, French, and Japanese. For every language, there are three domains: Book, Dvd, and Music. Each domain has 2,000 reviews for training and 2,000 for test, with 1,000 positive and 1,000 negative reviews in each set. 6,000 additional reviews form the unlabeled set for each domain. The source domains are only selected from the English corpus. 

Table~\ref{Tab:dataset} shows the data split. To construct the unlabeled set for the target domain, we use reviews from the test set as the unlabeled data in MonoAmazon following \citet{DBLP:conf/emnlp/HeLND18}. For MultiAmazon, reviews from the training set and original unlabeled set both from the target domain are combined. 

We also evaluate our model on the benchmark dataset of~\cite{DBLP:conf/acl/BlitzerDP07}. The results are presented in Appendix \ref{sec:benchmark}.

\subsection{Experimental Setup}
\noindent{\textbf{Model Settings.}} To enable cross-language transfer, we use XLM-R\footnote{\tiny{\url{https://github.com/pytorch/fairseq/tree/master/examples/xlmr}}}~\cite{conneau2019unsupervised} which has 25 layers as the pre-trained language model. The dimension of its token embeddings is 1024 which is mapped into 256 by the FAM. Based on one transfer result, the last 10-layer features are used in FAM. $\lambda$ for intra-class loss is set as 1 and 2 for MonoAmazon and MultiAmazon respectively. 
We set the size of negative sample set as 10 and we perform Fd training only in the target domain. $\tau$ for attention mechanism in Eq.~\ref{eq:tau} is set as 0.3. 
In the training process, we gradually increase the number of retained pseudo labels for self-training, in which we increase the number by 100 for MonoAmazon and 300 for MultiAmazon every epoch. 
$\alpha$ for $\mathcal{L}_{pred}^{T'}$ is the linear and quadratic function of epoch for MonoAmazon and MultiAmazon respectively.  
More details of the experimental settings are in Appendix~\ref{sec:setup}.

\noindent{\textbf{Baselines.}} Since we are interested in adapting features of PrLMs without tuning, we mainly set up the baselines that use the features from XLM-R by freezing XLM-R. 
Trained on the source domain, \textbf{xlmr-1} directly tests on the target without domain adaptation and it only uses the last-layer features of XLM-R.
\textbf{xlmr-10} is the same as xlmr-1, except that it uses the multi-layer representation of XLM-R with last 10-layer features. \textbf{KL} \cite{DBLP:conf/ijcai/ZhuangCLPH15} uses the balanced Kullback-Leibler divergence loss to decrease the domain discrepancy for domain adaptation. \textbf{MMD} adopts the Maximum Mean Discrepancy loss~\cite{DBLP:journals/jmlr/GrettonBRSS12} in which Gaussian Kernel is implemented. \textbf{Adv}~\cite{adv,chen2018adversarial} adversarially trains a domain classifier to learn domain-invariant features by reversing the gradients from the domain classifier following~\citet{adv}. \textbf{p} is our self-training method introduced in $\S$\ref{sec:self-train}. \textbf{p+CFd} is our full model that uses CFd to enhance the robustness of self-training.  \textbf{DAS}~\cite{DBLP:conf/emnlp/HeLND18} uses semi-supervised learning. 
\textbf{CLDFA}~\cite{DBLP:conf/acl/XuY17} is a cross-lingual baseline which uses cross-lingual resources.
\textbf{MAN-MoE}~\cite{DBLP:conf/acl/ChenAHWC19} studies multi-lingual transfer which has multiple languages in the source domain. MoE learns to focus on more transferable source domains for adaptation. xlmr-10, KL, MMD, Adv, p, and p+CFd are all based on the multi-layer representations with last 10-layer features.  For KL, MMD, and Adv, to minimize domain discrepancy, we use an unlabeled set of the same size in the source domain as the target domain. 

\textbf{xlmr-tuning}\footnote{Because of the limited computing resources, we cannot fine-tune XLM-R with unlabeled target data using LM loss.} first fine-tunes XLM-R with source labeled data using the representation from the final layer [CLS] and being fed to the classifier~\cite{bert}, then tests on the target. By setting up this baseline, we want to see how well the feature-based approach works. 

More detailed baseline settings can be found in Appendix~\ref{sec:baseline_setup}.

\begin{table*}[t]
\setlength{\abovecaptionskip}{0.2cm}
\setlength{\belowcaptionskip}{-0.2cm}
\centering
\setlength{\tabcolsep}{1mm}{\resizebox{\textwidth}{!}{%
\begin{tabular}{lccccccccccccc}
\toprule
\sc{Method}                & E$\to$BK       & BT$\to$BK      & M$\to$BK       & BK$\to$E       & BT$\to$E       & M$\to$E        & BK$\to$BT      & E$\to$BT       & M$\to$BT       & BK$\to$M       & E$\to$M        & BT$\to$M       & Ave.           \\
\hline
xlmr-10    & 70.41           & 67.80               & 70.83              & 56.47              & 70.65              & 64.74             & 61.30               & 71.57              & 65.38              & 63.33              & 67.69             & 64.47              & 65.16                                 \\ 

p                          & 70.90          & 71.38          & 72.18          & 64.00          & 70.41          & 67.01          & 67.48          & 71.67          & 70.71          & 67.16          & 67.92          & 69.77          & 69.21          \\  \cdashline{1-14}

\textbf{p+CFd}      & \textbf{75.25} & \textbf{74.70}     & \textbf{75.08}    & \textbf{70.19}    & \textbf{72.00}    & \textbf{68.96}   & \textbf{71.63}               & 73.73    & 70.05              & \textbf{70.86}    & \textbf{69.80}   & \textbf{70.46}    & {\color[HTML]{000000} \textbf{71.89}} \\
{p+C~(w/o Fd)} & 73.16          & 73.59          & 74.80          & 68.72          & 71.11          & 68.15          & 69.80          & \textbf{74.02} & \textbf{71.03} & 66.78          & 69.22          & 68.93          & 70.78          \\ 
{p+Fd~(w/o C)}  & 71.61          & 71.10          & 72.39          & 67.14          & 71.23          & 67.38          & 69.41          & 73.04          & 70.80          & 68.84          & 68.14          & 68.97          & 70.00          \\
{CFd~(w/o p)}        & 70.08           & 72.37               & 71.30              & 66.72              & 70.57              & 64.21             & 68.32               & 72.38              & 69.27              & 68.23              & 66.12             & 68.37              & 69.00                                 \\ 

{\color[HTML]{000000} Fd~(w/o p+C)} & {\color[HTML]{000000} 68.16}                        & {\color[HTML]{000000} {69.55}}                & {\color[HTML]{000000} 70.18}                        & {\color[HTML]{000000} {66.59}}               & {\color[HTML]{000000} 71.02}               & {\color[HTML]{000000} 63.92}                       & {\color[HTML]{000000} {69.18}}                & {\color[HTML]{000000} {72.10}}                & {\color[HTML]{000000} {67.77}}               & {\color[HTML]{000000} {69.73}}               & {\color[HTML]{000000} 65.71}                       & {\color[HTML]{000000} {66.13}}               & {\color[HTML]{000000} {68.34}}           \\
\toprule
\end{tabular}}%
}
\caption{The classification accuracy~(\%) results of p+CFd and its ablations on MonoAmazon.}
\label{Tab:ablation}
\end{table*}

\subsection{Main Results}\label{sec:main-results}

We conduct experiments in cross-domain (CD), cross-language (CL), and both cross-language and cross-domain (CLCD) settings. Results of CD are evaluated on MonoAmazon~(Table~\ref{Tab:cross-domain}) and results of CL and CLCD are on MultiAmazon~(Table~\ref{Tab:cross-lang}). For CL, English is set as the source language. The domains in the source and target languages are the same, i.e., When German\&book is the target, the source will be English\&book. For CLCD, the sources are also only from English. For example, when the target is German\&book, the source language is English and the source domain is dvd or music, in which two sources are set up:  English\&dvd and English\&music, and the two adaptation results are averaged for German\&book. 

We have the following findings from Table~\ref{Tab:cross-domain} and \ref{Tab:cross-lang} based on the overall average scores. \textbf{xlmr-10 vs. xlmr-tuning:} xlmr-10 is slightly better than xlmr-tuning which demonstrates the effectiveness of the feature-based approach. \textbf{xlmr-1 vs. xlmr-10:} xlmr-10 is much better than xlmr-1 which means our multi-layer representation of XLM-R is much more transferable than the last-layer feature. \textbf{xlmr-10 vs. p:} p is consistently better than xlmr-10 which shows our self-training method is effective. \textbf{p vs. p+CFd:} After using CFd, p can be consistently improved and p+CFd achieves the best performance among all the methods, which shows the effectiveness of CFd.

\subsection{Further Analysis}\label{sec:further-analysis}
\noindent{\textbf{Ablation Study.}} We conduct the ablation experiments to see the contributions of feature self-distillation~(Fd) and class information~(C), which are evaluated on MonoAmazon based on last 10-layer features. By ablating p+CFd, we have four baselines of p+C~(w/o Fd), p+Fd~(w/o C), CFd~(w/o p) and Fd~(w/o p+C). From the results in Table~\ref{Tab:ablation}, p+Fd and p+C perform worse than p+CFd but still better than p, so feature self-distillation and class information both contribute to the improvements of p. 
Also, by removing the effects of p, CFd and Fd substantially outperform xlmr-10, which means CFd and Fd are both effective for domain adaptation, independent of the self-training method. 

\begin{figure}[t]
\setlength{\abovecaptionskip}{-0.1cm}
\setlength{\belowcaptionskip}{-0.2cm}
\begin{center}
\includegraphics[width=\columnwidth]{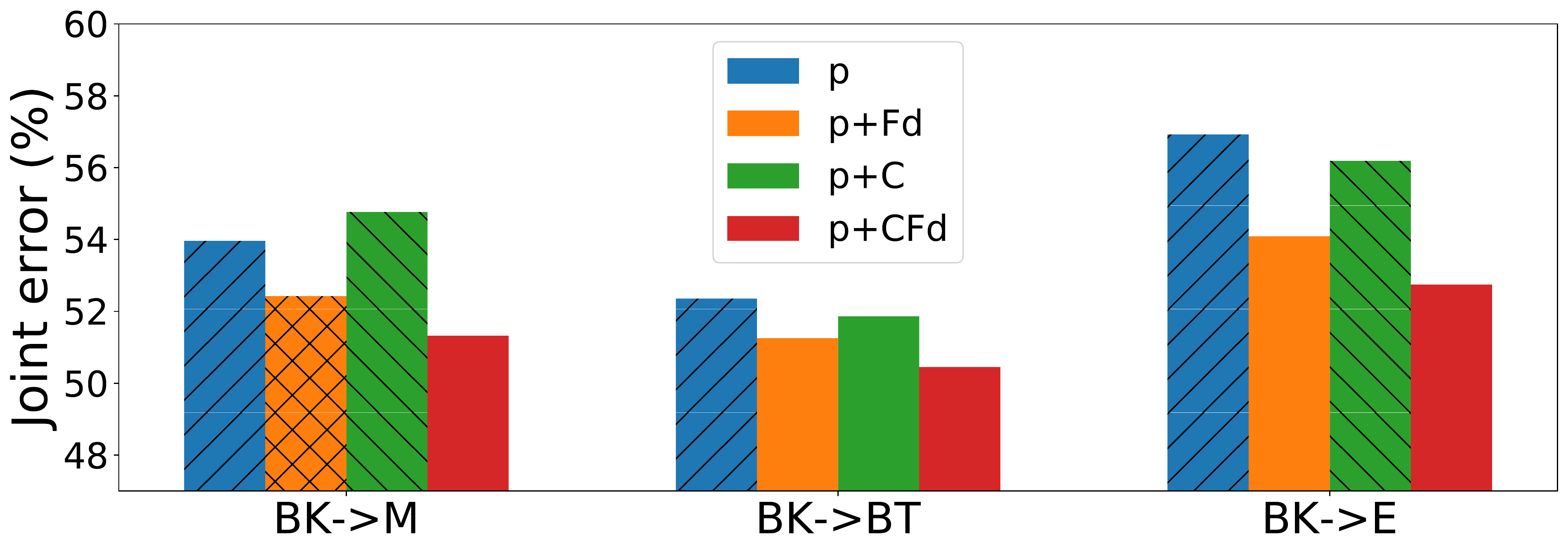}
\end{center}
\caption{The errors of ideal joint hypothesis tested on MonoAmazon.}
\label{fig:joint_error}
\end{figure}

\noindent{\textbf{Joint errors.}} Here we study why CFd can enhance self-training and provide empirical results to demonstrate the theoretical understanding in $\S$\ref{sec:analysis}. By testing on MonoAmazon based on last 10-layer features, Figure~\ref{fig:joint_error} presents the joint error results. For example, to find $h^*$ in Eq.~\ref{eq:joint_error} for baseline p, following \citet{DBLP:conf/icml/LiuLWJ19}, we train a classifier using the combined source and target labeled data based on the fixed FAM trained by p. We note that p+Fd and p+C can achieve lower joint errors compared to p, and p+CFd has the best performance, which is consistent with our analysis in $\S$\ref{sec:analysis}. 

\begin{table}[]
\setlength{\abovecaptionskip}{0.2cm}
\setlength{\belowcaptionskip}{-0.2cm}
\centering
\resizebox{\columnwidth}{!}{
\setlength{\tabcolsep}{0.2mm}{\begin{tabular}{lccccccc}
\toprule
\sc Method & \multicolumn{1}{l}{M$\to$BK} & \multicolumn{1}{c}{BK$\to$E} & \multicolumn{1}{c}{M$\to$E} & \multicolumn{1}{c}{BK$\to$BT} & \multicolumn{1}{c}{M$\to$BT} & \multicolumn{1}{c}{BK$\to$M} & \multicolumn{1}{c}{Ave.}     \\ \hline

Super          & 77.54                        & 74.56                        & 74.08                       & 75.66                         & 75.48                        & 72.88                        & 75.03                        \\ 
Fd             & 76.34                        & 72.44                        & 72.44                       & 74.35                         & 73.15                        & 74.12                        & 73.81                        \\ \toprule

\end{tabular}}
}
\caption{The classification accuracy~(\%) results of in-domain test evaluated on MonoAmazon.}
\label{Tab:in-domain-test}
\end{table}

\noindent{\textbf{Effects of Feature Self-distillation.}} We conduct an in-domain test to verify that Fd learns discriminative features from the PrLM. 
We build a sentiment classification model with in-domain data based on the last 10-layer features. From the same domain in MonoAmazon, we select 4,000 labeled pairs for training, 1,000 for validation, and 1,000 for test. 
We first pre-train the FAM by Fd using the entire 6,000 raw texts, then we freeze FAM and train a classifier with the training data with features out of FAM. We compare the results with the baseline that directly trains the FAM and classifier with training set~(Super). From the results in Table~\ref{Tab:in-domain-test}, the performances of Fd are very close to Super, showing that the features out of FAM after Fd training are discriminative.

\noindent{\textbf{Effects of Class Information.}} Table~\ref{Tab:class} presents the average intra-class loss in the training process. By exploring class information, the intra-class loss can be dramatically minimized and accordingly the transfer performances are improved.

\begin{table}[]
\setlength{\abovecaptionskip}{0.2cm}
\setlength{\belowcaptionskip}{-0.2cm}
\centering
\resizebox{\columnwidth}{!}{
\setlength{\tabcolsep}{4.mm}{\begin{tabular}{lcccc}
\toprule

     BK$\to$M  & p      & p+C   & p+Fd   & p+CFd \\ \hline
$\mathcal{L}_{intra\_class}$ & {966.38} & \textbf{11.00} & {327.72} & \textbf{12.17} \\ 
Acc. (\%)                    & {66.59}  & \textbf{68.88} & {70.16}  & \textbf{70.95} \\ \toprule

\end{tabular}}
}
\caption{Effects of class information tested on MonoAmazon with last 10-layer features.}
\label{Tab:class}
\end{table}

\begin{figure}[t]
\setlength{\abovecaptionskip}{-0.1cm}
\setlength{\belowcaptionskip}{-0.2cm}
\begin{center}
\includegraphics[width=\columnwidth]{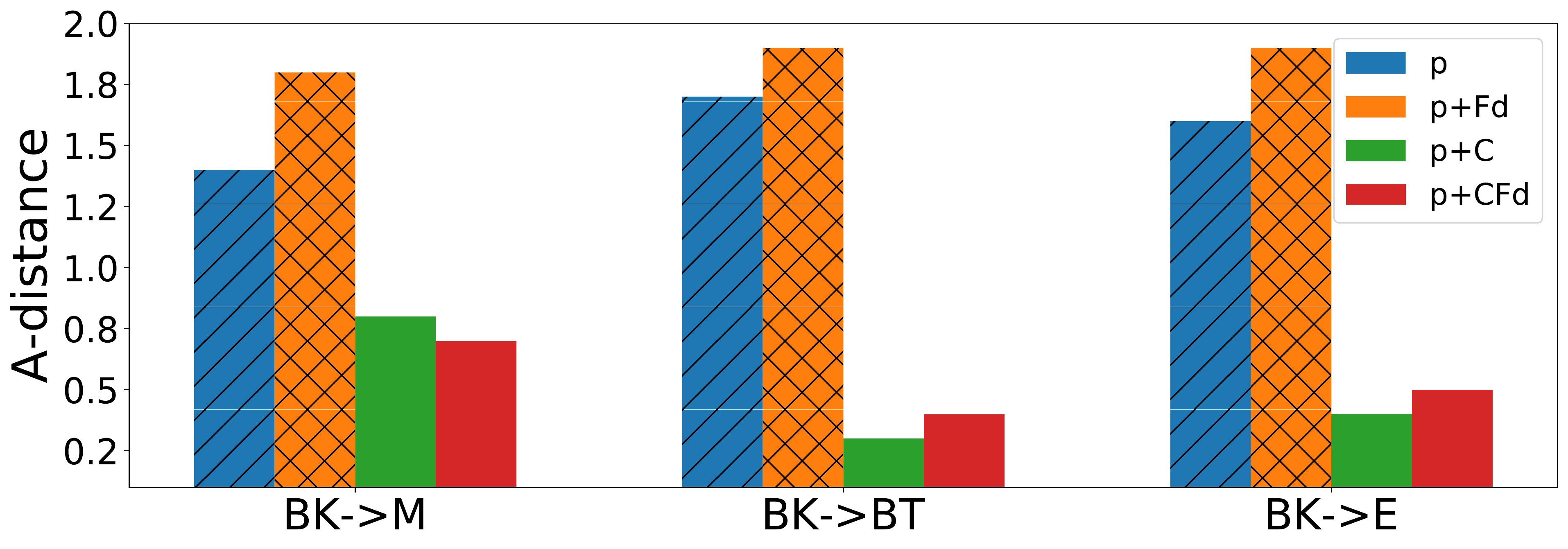}
\end{center}
\caption{The $\mathcal{A}$-distance tested on MonoAmazon.}
\label{fig:a_distance}
\end{figure}

\noindent{\textbf{$\mathcal{A}$-distance}.} As an indicator of domain discrepancy, following \citet{tri-net}, we calculate the $\mathcal{A}$-distance based on the last 10-layer features out of FAM trained by method of p or others, and train a classifier to classify the source and target domain data. $d_{\mathcal{A}}$ is equal to $2 (1-\delta)$ and $\delta$ is the domain classification error. From Figure~\ref{fig:a_distance}, p+C and p+CFd have much smaller $\mathcal{A}$-distance, which means that the intra-class loss reduces the domain discrepancy. p+Fd has larger $\mathcal{A}$-distance, probably because Fd learns domain-specific information from the target so the domain distance becomes larger. 

\noindent{\textbf{Effects of Attention Mechanism.}} We further show whether combining the intermediate-layer features can enhance adaptation. In Table~\ref{Tab:att}, one layer means only using one-layer features for transfer and the results are obtained by using the feature from the most transferable layer. We introduce the attention mechanism to combine the last $N$-layer features. We demonstrate that using last 10-layer features with attention can achieve better performances. AVE that averages the last $N$-layer features cannot improve the performance, since it lacks the ability to focus more on effective features. 
\begin{table}[]
\setlength{\abovecaptionskip}{0.2cm}
\setlength{\belowcaptionskip}{-0.2cm}
\centering
\resizebox{\columnwidth}{!}{
\setlength{\tabcolsep}{3mm}{\begin{tabular}{lccccc}
\toprule
\multirow{2}{*}{\sc Method} & \multirow{2}{*}{One layer} & \multicolumn{2}{c}{last-10} & \multicolumn{2}{c}{last-20} \\ \cline{3-6} 
                            &                                   & AVE      & ATT              & AVE      & ATT              \\ \hline
BK$\to$M                    & 69.51                             & 69.20    & \textbf{70.07}   & 66.62    & 69.31            \\ 
BK$\to$BT                   & 69.27                             & 67.62    & \textbf{69.34}   & 64.16    & 69.02            \\ 
BK$\to$E                    & 66.35                             & 64.62    & 66.71            & 62.90    & \textbf{67.08}   \\ \toprule
\end{tabular}}
}
\caption{Study of our attention mechanism based on Fd baseline and tested on MonoAmazon.}
\label{Tab:att}
\end{table}

We also study how the size of negative sample set affects feature distillation and the effects of sharpen on attention mechanism. The analysis is included in Appendix~\ref{sec:further}.

\section{Conclusion}
In this paper, we study how to adapt the features from the pre-trained language models without tuning. We specifically study unsupervised domain adaptation of PrLMs, where we transfer the models trained in labeled source domain to the unlabeled target domain based on PrLM features. We build our adaptation method based on self-training. To enhance the robustness of self-training, we present the method of class-aware feature self-distillation to learn discriminative features.
Experiments on sentiment analysis in cross-language and cross-domain settings demonstrate the effectiveness of our method.

\section*{Acknowledgments}
This work was supported by Alibaba Group through the Alibaba Innovative Research (AIR) Program.

\bibliographystyle{acl_natbib}
\bibliography{emnlp2020}


\appendix



\section{Experimental Settings}\label{sec:setup}
\subsection{Datasets} We obtain the datasets from \citet{DBLP:conf/emnlp/HeLND18} which can be downloaded online\footnote{\url{https://github.com/ruidan/DAS}}. Then we follow \citet{DBLP:conf/emnlp/HeLND18} to pre-process the datasets which only involves splitting the data into training, validation, and test sets. 

\begin{table*}[t]
\centering
\resizebox{\textwidth}{!}{%
\begin{tabular}{l|c|c|c}
\toprule
\textbf{parameter}       & MonoAmazon              & MultiAmazon                 & Benchmark                   \\ \hline
learning rate               & 0.0001                  & 0.0005                      & 0.0005                      \\ 
$\lambda$                   & 1                       & 2                           & 2                           \\ 
$\alpha$                    & linear function of epoch & quadratic function of epoch & quadratic function of epoch \\ 
Max epoch                   & 35                      & 20                          & 20                          \\ 
Size of negative sample set & 10                      & 10                          & 10                          \\ \bottomrule
\end{tabular}
}
\caption{Hyper-parameter settings for main experiments.}
\end{table*}

\begin{table}[]
\setlength{\abovecaptionskip}{0.2cm}
\setlength{\belowcaptionskip}{-0.3cm}
\centering
\setlength{\tabcolsep}{1mm}{\resizebox{8cm}{!}{
\begin{tabular}{lccccc}
\toprule[0.5pt]
  \sc Data          & train~(\emph{S}) & valid~(\emph{S}) & test~(\emph{T}) & unlabeled~(\emph{T}) & $|C|$ \\ \hline
Benchmark      & 1,600     & 400      & 400    & 6,000     & 2   \\  \toprule[0.5pt]
\end{tabular}}
}
\caption{The data split for training, validation, test, and unlabeled set on Benchmark. $|C|$ is the number of classes.}
\label{Tab:dataset-mono}
\end{table}

\begin{table*}[!]
\setlength{\abovecaptionskip}{0.2cm}
\setlength{\belowcaptionskip}{-0.5cm}
\centering
\resizebox{\textwidth}{!}{
\setlength{\tabcolsep}{0.8mm}{\begin{tabular}{l|lllllllllllll}
\toprule

\textbf{Benchmark} & D$\to$B        & K$\to$B        & E$\to$B        & B$\to$D        & K$\to$D        & E$\to$D         & B$\to$K         & D$\to$K         & E$\to$K         & B$\to$E         & D$\to$E        & K$\to$E         & Ave.                         \\ \hline
AsyTri & 73.20 & 72.50 & 73.20 & 80.70 & 74.90 & 72.90 & 82.50 & 82.50 & 86.90 & 79.80 & 77.00 & 84.60 & 78.39 \\
DAS & 82.05 & 80.05  & 80.00 & 82.75 & 81.40 & 80.15 & 82.25 & 81.50 & 84.85 & 81.15 & 81.55 & 85.80 & 81.96 \\ \cdashline{1-14}
xlmr-1                    & 88.50          & 78.45          & 82.50          & 85.25          & 80.55          & 81.80           & 84.50           & 81.15           & 88.45           & 81.25           & 79.35          & 90.05           & 83.48                        \\ 
xlmr-10                   & 91.30$_{1.0}$          & 87.95$_{1.0}$          & 87.95$_{0.3}$          & \textbf{87.90}$_{0.5}$ & 87.05$_{0.6}$          & 86.85$_{0.4}$           & 90.45$_{1.0}$           & 87.55$_{1.5}$           & 92.30$_{0.7}$           & 88.90$_{0.5}$           & 89.05$_{1.7}$          & 91.60$_{0.3}$           & 89.07                        \\ 
KL                        & \textbf{91.50}$_{0.8}$ & 88.95$_{0.6}$          & 88.05$_{0.5}$          & 87.20$_{0.6}$          & \textbf{87.85}$_{0.5}$ & 87.30$_{0.6}$           & 90.00$_{1.0}$           & 91.15$_{0.3}$           & 92.70$_{0.4}$           & 89.70$_{0.6}$           & 90.65$_{0.2}$          & 91.35$_{1.0}$           & {\color[HTML]{000000} 89.70} \\ 
MMD                       & 91.75$_{0.5}$          & 88.65$_{1.1}$          & 87.55$_{0.9}$          & 87.05$_{0.7}$          & 86.45$_{0.3}$          & 86.50$_{0.6}$           & 90.05$_{0.3}$           & 90.70$_{0.5}$           & 92.30$_{0.3}$           & 90.15$_{0.3}$           & 91.50$_{0.6}$          & 91.65$_{0.7}$           & {\color[HTML]{000000} 89.53} \\ 
Adv                       & 91.40$_{0.8}$          & 88.10$_{0.4}$          & 88.15$_{0.4}$          & 87.70$_{1.0}$          & 87.35$_{0.8}$          & 86.65$_{0.3}$           & 90.65$_{0.5}$           & 87.55$_{1.5}$           & 92.25$_{0.2}$           & 89.25$_{0.5}$           & 89.80$_{1.3}$          & 91.60$_{0.6}$           & {\color[HTML]{000000} 89.20} \\ \hline
p                         & 91.40$_{0.3}$          & 89.50$_{0.4}$          & 88.20$_{0.6}$          & 87.40$_{0.3}$          & 87.15$_{0.3}$          & 87.05$_{0.9}$           & 90.00$_{0.6}$           & 87.55$_{1.7}$           & 92.60$_{0.3}$           & 88.85$_{0.2}$           & 89.65$_{1.9}$          & 91.85$_{0.4}$           & 89.27                        \\

\textbf{p+CFd}            & \textbf{91.50}$_{0.4}$ & \textbf{89.75}$_{0.8}$ & \textbf{88.65}$_{0.4}$ & 87.65$_{0.1}$          & 87.80$_{0.4}$          & \textbf{88.20}$_{0.4}$ & \textbf{92.45}$_{0.6}$ & \textbf{92.45}$_{0.2}$ & \textbf{93.60}$_{0.5}$ & \textbf{91.30}$_{0.2}$ & \textbf{91.55}$_{0.3}$ & \textbf{92.60}$_{0.5}$ & \textbf{90.63}               \\ \bottomrule
\end{tabular}}}
\caption{The cross-domain classification accuracy~(\%) results on Benchmark. Models are evaluated by 5 random runs. We report the mean and standard deviation results. The best task performance is boldfaced. 
Results of DAS and AsyTri are taken from \citet{DBLP:conf/emnlp/HeLND18} and \citet{tri-net} respectively. \textbf{AsyTri}~\cite{tri-net} is a self-training baseline with tri-training.}
\label{Tab:cross-domain-Benchmark}
\end{table*}

\subsection{Model Configuration}
For MonoAmazon, the learning rate is 0.0001, and the batch size is 50 for classifier training and MI learning. We run 35 times for each baseline except xlmr-1 and xlmr-10 which are run 20 times and the batch size is 100. In epoch 0, we set to retain the top 950 high-confidence predictions for self-training and we increase the number of retained data by 100 every epoch. $\lambda$ for Fd training is 1. 

For MultiAmazon and Benchmark, the learning rate is 0.0005. The batch size for classifier learning is 50 and for MI training is 200. The training epoch is 20. $\lambda$ for MultiAmazon and Benchmark is 2. In epoch 0, we set to retain the top 1000 high-confidence predictions for self-training. We increase by 150 retained samples every epoch for Benchmark, and by 300 for MultiAmazon.

$\alpha$ for $\mathcal{L}_{pred}^{T'}$ is the linear function of epoch for MonoAmazon,  and the quadratic function for MultiAmazon and Benchmark. Adam~\cite{DBLP:journals/corr/KingmaB14} is used for model training. In the training process, if the validation performance does not improve after 10 consecutive epochs, the learning rate will be halved.

For all the datasets, the size of negative sample set is set as 10. $\tau$ for attention mechanism is set as 0.3, tuned from \{0.1, 0.3, 0.5, 0.8, 1.0\}.

\subsection{Settings for Baselines}\label{sec:baseline_setup}
\noindent{\textbf{KL}.} The KL-divergence loss~\cite{DBLP:conf/ijcai/ZhuangCLPH15} is defined as:
\begin{equation}
     \text{KL} = D_{KL}(\mathbf{\xi}_s || \mathbf{\xi}_t) + D_{KL}(\mathbf{\xi}_t || \mathbf{\xi}_s)
\end{equation}
where 
\begin{equation}
\begin{split}
    \xi_s' &= \frac{1}{n}\sum\limits_{i=1}^n \mathbf{z}_s^{i} \ \ \ \ \xi_s = softmax(\xi_s') \\
    \xi_t' &= \frac{1}{n}\sum\limits_{i=1}^n \mathbf{z}_t^{i} \ \ \ \ \xi_t = softmax(\xi_t')
\end{split}
\end{equation}
in which $n$ is the batch size. We set the weight of KL loss as 500, tuned from \{100, 500, 1000, 5000\}.

\noindent{\textbf{MMD.}} We use the Gaussian kernel to implement the MMD loss~\cite{DBLP:journals/jmlr/GrettonBRSS12}. The kernel number is 5. The weight for MMD loss is set to 1, tuned from \{1, 0.1, 0.5\}

\noindent{\textbf{Adv.}} We follow \citet{adv} to reverse the gradients from the domain classifier. We set the learning rate for Adv to be the same as the baselines, but set the weight for domain classifier as 0.01, tuned from \{1, 0.1, 0.01, 0.001\}.

\noindent{\textbf{xlmr-tuning.}} The fine-tuning baseline uses the first [CLS] token as the document representation. The learning rate is 1e-5 and the batch size for gradient update is 32. The fine-tuning models generally overfit the training data in 5 epochs.

\section{Results on Benchmark}\label{sec:benchmark}
\noindent{\textbf{Benchmark.}} This is a benchmark dataset for domain adaptation~\cite{DBLP:conf/acl/BlitzerDP07}, whose reviews are also in English. Four domains are included: Book~(B), DVDs~(D), Electronics~(E), and Kitchen~(K). Each domain has 1,000 positive and 1,000 negative reviews. 
Following~\citet{DBLP:conf/emnlp/HeLND18}, there are 4,000 unlabeled reviews for each domain. Table~\ref{Tab:dataset-mono} summarizes the data split when training on Benchmark. The unlabeled set is the combination of the training set and the original unlabeled set. Table~\ref{Tab:cross-domain-Benchmark} shows the results on Benchmark.

\section{Further Analysis}\label{sec:further}
\begin{figure*}[]
\setlength{\abovecaptionskip}{-0.1cm}
\setlength{\belowcaptionskip}{-0.5cm}
\begin{center}
\includegraphics[width=10cm]{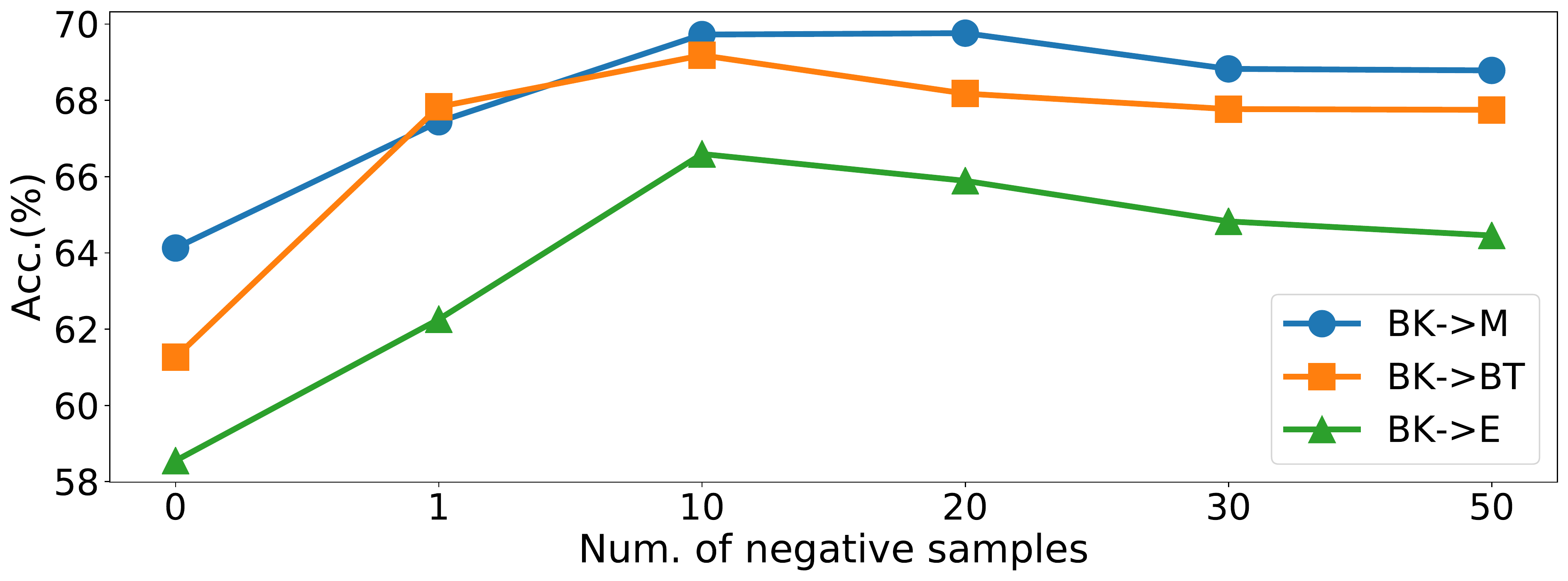}
\end{center}
\caption{The effects of the negative sample set size for feature self-distillation. Method is xlmr-10+Fd which is evaluated on MonoAmazon.}
\label{fig:neg_n}
\end{figure*}

\noindent{\textbf{Size of Negative Sample Set.}} We study how the size of negative sample set will affect Fd training. The results are shown in Fig.~\ref{fig:neg_n}. The method used is xlmr-10+Fd. We find that using a size that is too small or too big is not a good strategy for Fd learning. Size of 10 is a good option for Fd learning. 

\begin{figure*}[]
\setlength{\abovecaptionskip}{-0.2cm}
\setlength{\belowcaptionskip}{-0.2cm}
\begin{center}
\includegraphics[width=10cm]{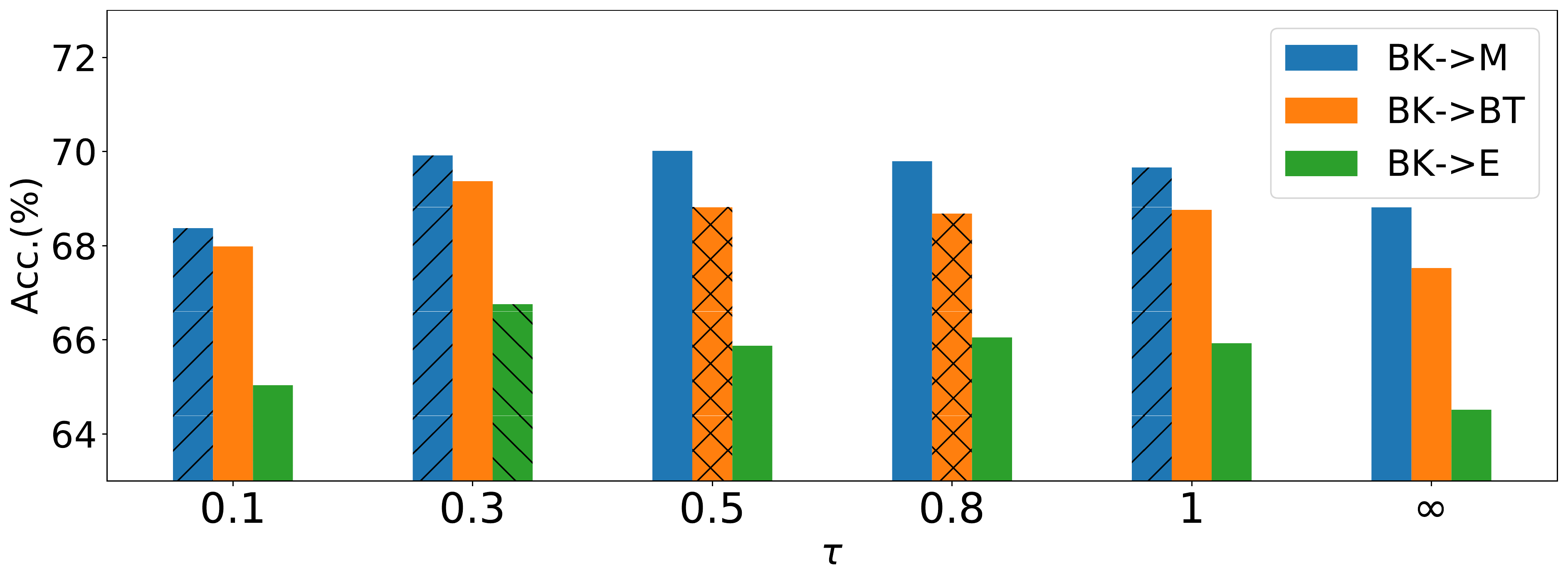}
\end{center}
\caption{Effects of sharpen on MonoAmazon with method of xlmr-10+Fd.}
\label{fig:sharpen}
\end{figure*}

\noindent{\textbf{Effects of Sharpen on Attention Mechanism.}} In Fig.~\ref{fig:sharpen}, we show the effects of sharpen mechanism in our attention method which demonstrates that when not using sharpen~($\tau$ is $\infty$), the performance will drop and $\tau$ set as 0.3 is a good option for our attention method. 

\begin{table*}[]
\centering
\resizebox{\textwidth}{!}{%
\begin{tabular}{lcccccccccc}
\toprule
\multicolumn{11}{c}{English $\to$ German}                               \\ \hline
S              & book                              & dvd                               & music                             & book                              & dvd                               & music                             & book                              & dvd                               & music                             &                           \\ 
T              & book                              & book                              & book                              & dvd                               & dvd                               & dvd                               & music                             & music                             & music                             & Ave.                       \\ \hline
xlmr-tuning    & \multicolumn{1}{l}{91.03$_{0.3}$} & \multicolumn{1}{l}{91.03$_{0.5}$} & \multicolumn{1}{l}{90.65$_{0.3}$} & \multicolumn{1}{l}{88.47$_{0.4}$} & \multicolumn{1}{l}{88.02$_{0.6}$} & \multicolumn{1}{l}{88.48$_{0.3}$} & \multicolumn{1}{l}{89.75$_{0.4}$} & \multicolumn{1}{l}{89.75$_{0.7}$} & \multicolumn{1}{l}{90.13$_{0.2}$} & \multicolumn{1}{l}{89.70} \\ 
xlmr-1         & 73.69                             & 62.08                             & 86.12                             & 68.03                             & 69.86                             & 86.28                             & 66.4                              & 66.63                             & 87.34                             & 74.05                     \\ 
xlmr-10        & 93.15$_{0.8}$                     & 93.79$_{0.6}$                     & 88.20$_{1.4}$                     & 87.22$_{1.1}$                     & 89.59$_{1.2}$                     & 84.68$_{1.3}$                     & 92.33$_{1.5}$                     & 92.63$_{0.5}$                     & 92.26$_{0.6}$                     & 90.43                     \\ 
KL             & \textbf{93.99}$_{0.4}$                     & 93.99$_{0.1}$                     & 92.49$_{0.2}$                     & 90.81$_{0.4}$                     & 91.12$_{0.4}$                     & 89.96$_{0.3}$                     & 93.13$_{0.1}$                     & 92.87$_{0.4}$                     & \textbf{93.89$_{0.2}$}            & 92.47                     \\ 
MMD            & 93.97$_{0.1}$                     & 93.81$_{0.4}$                     & 93.07$_{0.1}$                     & 90.89$_{0.3}$                     & 90.77$_{0.8}$                     & 90.10$_{0.2}$                     & 92.92$_{0.1}$                     & 92.23$_{0.5}$                     & 93.53$_{0.4}$                     & 92.37                     \\ 
Adv            & 93.27$_{0.4}$                     & 94.11$_{0.6}$                     & 91.41$_{0.3}$                     & 90.39$_{1.2}$                     & 89.78$_{0.6}$                     & 87.14$_{0.4}$                     & 92.99$_{0.2}$                     & 92.61$_{0.4}$                     & 92.53$_{0.6}$                     & 91.58                     \\ \hline
p              & 92.99$_{1.0}$                     & 93.89$_{0.5}$                     & 92.33$_{0.1}$                     & 87.83$_{1.5}$                     & 89.33$_{0.6}$                     & 89.03$_{0.6}$                     & 92.97$_{0.3}$                     & 92.70$_{0.3}$                     & 93.82$_{0.3}$                     & 91.65                     \\ 
\textbf{p+CFd} & 93.95$_{0.2}$                     & \textbf{94.83$_{0.1}$}            & \textbf{93.74$_{0.2}$}            & \textbf{91.03$_{0.1}$}            & \textbf{91.69$_{0.3}$}            & \textbf{90.42$_{0.4}$}            & \textbf{93.59$_{0.3}$}            & \textbf{93.65$_{0.3}$}            & \textbf{93.89$_{0.2}$}            & \textbf{92.98}            \\ \bottomrule

\\
\toprule
\multicolumn{11}{c}{English $\to$ French}           \\                                                                                                                                                                \hline
S              & book                   & dvd           & music                  & book          & dvd                    & music                  & book                   & dvd                    & music                  &                      \\
T              & book                   & book          & book                   & dvd           & dvd                    & dvd                    & music                  & music                  & music                  & Ave.                  \\ \hline
xlmr-tuning    & 92.12$_{0.5}$          & 90.70$_{0.3}$ & 89.88$_{0.9}$          & 90.70$_{0.4}$ & 91.17$_{0.3}$          & 90.38$_{0.5}$          & 90.17$_{0.5}$          & 89.13$_{0.5}$          & 89.58$_{0.8}$          & 90.43                \\ 
xlmr-1         & 91.26                  & 89.44         & 86.46                  & 89.33         & 91.13                  & 86.67                  & 87.18                  & 89.11                  & 88.37                  & 88.77                \\ 
xlmr-10        & 93.79$_{0.4}$          & 92.67$_{0.7}$ & 87.67$_{0.8}$          & 93.21$_{0.2}$ & 93.28$_{0.4}$          & 87.37$_{1.8}$          & 92.86$_{0.4}$          & 92.45$_{0.5}$          & 92.23$_{0.6}$          & 91.73                \\ 
KL             & 93.91$_{0.1}$          & \textbf{93.59}$_{0.2}$ & 90.37$_{0.2}$          & 92.96$_{0.3}$ & 93.31$_{0.3}$          & 92.09$_{0.2}$          & 92.51$_{0.7}$          & 93.11$_{0.1}$          & 92.39$_{0.2}$          & 92.69                \\ 
MMD            & 93.48$_{0.2}$          & 93.55$_{0.2}$ & 91.85$_{0.6}$          & 92.85$_{0.2}$ & 93.21$_{0.2}$          & 92.21$_{0.2}$          & 93.34$_{0.4}$          & 92.80$_{0.6}$          & 92.67$_{0.2}$          & 92.88                \\ 
Adv            & 93.70$_{0.4}$          & 93.42$_{0.3}$ & 89.73$_{0.7}$          & 93.14$_{0.5}$ & 93.03$_{0.4}$          & 90.26$_{0.6}$          & 92.43$_{0.6}$          & 92.85$_{0.3}$          & 92.28$_{0.3}$          & 92.32                \\ \hline
p              & 93.81$_{0.1}$          & 93.57$_{0.2}$ & 90.61$_{0.5}$          & 93.14$_{0.3}$ & 93.00$_{0.2}$          & 91.68$_{0.3}$          & 92.24$_{0.7}$          & 92.80$_{0.3}$          & 92.50$_{0.2}$          & 92.59                \\ 
\textbf{p+CFd} & \textbf{94.25$_{0.2}$} & 93.40$_{0.3}$ & \textbf{92.80$_{0.2}$} & 93.10$_{0.4}$ & \textbf{93.79$_{0.1}$} & \textbf{92.51$_{0.1}$} & \textbf{93.33$_{0.6}$} & \textbf{93.91$_{0.2}$} & \textbf{93.39$_{0.1}$} & \textbf{93.39}       \\ \bottomrule
\\

\toprule

\multicolumn{11}{c}{English $\to$ Japanese}                                                                                                                                                                                                                   \\ \hline
S           & book                   & dvd                    & music                  & book                   & dvd                    & music                  & book                   & dvd                    & music                  &                \\ 
T           & book                   & book                   & book                   & dvd                    & dvd                    & dvd                    & music                  & music                  & music                  & Ave.            \\ \hline
xlmr-tuning & 87.52$_{0.5}$          & 85.90$_{0.6}$          & 85.90$_{0.4}$          & 86.13$_{0.3}$          & 87.12$_{0.4}$          & 85.90$_{0.4}$          & 88.18$_{0.2}$          & 87.52$_{0.4}$          & 88.52$_{0.7}$          & 86.96          \\ 
xlmr-1      & 70.96                  & 68.18                  & 84.73                  & 64.96                  & 71.2                   & 85.43                  & 61.81                  & 70.04                  & 87.07                  & 73.82          \\ 
xlmr-10     & 87.13$_{1.1}$          & 87.52$_{0.6}$          & 83.81$_{1.6}$          & 87.88$_{1.0}$          & 88.63$_{0.1}$          & 83.49$_{2.3}$          & 88.94$_{0.3}$          & 86.83$_{1.4}$          & 88.05$_{0.5}$          & 86.92          \\ 
KL          & 88.60$_{0.1}$          & 87.53$_{0.4}$          & 85.76$_{0.5}$          & \textbf{88.88$_{0.3}$} & 88.82$_{0.2}$          & 87.53$_{0.2}$          & 88.80$_{0.4}$          & 88.41$_{0.2}$          & 88.12$_{0.2}$          & 88.05          \\ 
MMD         & 89.17$_{0.1}$          & 88.20$_{0.1}$          & 87.29$_{0.2}$          & 88.80$_{0.3}$          & \textbf{89.22$_{0.1}$} & 87.69$_{0.5}$          & 89.23$_{0.3}$          & 88.23$_{0.5}$          & 88.54$_{0.4}$          & 88.49          \\ 
Adv         & 88.22$_{0.8}$          & 87.72$_{0.3}$          & 86.04$_{0.5}$          & 88.64$_{0.5}$          & 88.68$_{0.1}$          & 87.57$_{0.4}$          & 88.17$_{1.3}$          & 87.89$_{0.3}$          & 88.34$_{0.2}$          & 87.92          \\ \hline
p           & 88.68$_{0.3}$          & 87.95$_{0.2}$          & 86.25$_{0.5}$          & 88.77$_{0.2}$          & 88.86$_{0.1}$          & 87.67$_{0.2}$          & 88.89$_{0.3}$          & 88.25$_{0.3}$          & 88.39$_{0.1}$          & 88.19          \\ 
\textbf{p+CFd}       & \textbf{89.41$_{0.2}$} & \textbf{88.78$_{0.2}$} & \textbf{89.08$_{0.1}$} & 88.77$_{0.5}$          & 88.68$_{0.1}$          & \textbf{89.22$_{0.3}$} & \textbf{89.83$_{0.2}$} & \textbf{88.98$_{0.2}$} & \textbf{89.54$_{0.3}$} & \textbf{89.14} \\ \bottomrule

\end{tabular}%
}
\caption{Full classification accuracy~(\%) results on MultiAmazon. Models are evaluated by 5 random runs except xlmr-tuning which is run for 3 times to save time. We report the mean and standard deviation results. The best task performance is boldfaced.}
\label{Tab:full-multiamazon}
\end{table*}

\section{Full Results on MultiAmazon}\label{sec:multiAmazon}
Table~\ref{Tab:full-multiamazon} shows the full results on MultiAmazon.

\end{document}